\documentclass[letterpaper, 10 pt, conference]{ieeeconf}  

\IEEEoverridecommandlockouts                              

\usepackage{graphics} 
\usepackage{epsfig} 
\usepackage{mathptmx} 
\usepackage{times} 
\usepackage{amsmath} 
\usepackage{amssymb}  
\usepackage{bm}
\usepackage{algorithm}
\usepackage{multirow}
\usepackage{verbatim}
\usepackage{graphicx}
\usepackage{hyperref}
\usepackage{textcomp}
\usepackage{cite}
\usepackage{subcaption}
\usepackage{lipsum}
\usepackage{mwe}
\usepackage[font=footnotesize]{caption}

\usepackage{titlesec}
\titlespacing{\section}{0pt}{*0.6}{*0.6}
\titlespacing{\subsection}{4pt}{*0.5}{*0.5}

\IEEEoverridecommandlockouts                   
\usepackage{booktabs} 
\title{\LARGE \bf
Virtual Barriers in Augmented Reality for Safe and Effective Human-Robot Cooperation in Manufacturing
}

\author{Khoa Cong Hoang, Wesley P. Chan, Steven Lay, Akansel Cosgun and Elizabeth Croft \\ Department of Electrical and Computer Systems Engineering, Monash University} 
\begin{document}

\maketitle
\thispagestyle{empty}
\pagestyle{empty}

\begin{abstract}
Safety is a fundamental requirement in any human-robot collaboration scenario. To ensure the safety of users for such scenarios, we propose a novel Virtual Barrier system facilitated by an augmented reality interface. Our system provides two kinds of Virtual Barriers to ensure safety: 1) a Virtual Person Barrier which encapsulates and follows the user to protect them from colliding with the robot, and 2) Virtual Obstacle Barriers which users can spawn to protect objects or regions that the robot should not enter. To enable effective human-robot collaboration, our system includes an intuitive robot programming interface utilizing speech commands and hand gestures, and features the capability of automatic path re-planning when potential collisions are detected as a result of a barrier intersecting the robot's planned path. We compared our novel system with a standard 2D display interface through a user study, where participants performed a task mimicking an industrial manufacturing procedure. Results show that our system increases the user's sense of safety and task efficiency, and makes the interaction more intuitive.

\end{abstract}

\section{INTRODUCTION}
        In any manufacturing process, safety of workers is paramount. Current robot-assisted manufacturing processes use light curtains or other physical barriers such as safety cages around robots to protect workers, and the robot operations are stopped when a worker enters the robot's vicinity. In such cases, a teach pendant or a 2D display-based interface is usually employed for programming and interacting with the robot. Recently, interest in bringing robots and humans workers together to collaborate with each other has been on the rise. In such scenarios, these standard methods for interacting with robots are not suitable, as they are unintuitive and inflexible, and their complexity often distracts the user from the main task, greatly affecting productivity and level of safety \cite{Guerin,BROGARDH,Heyer}. 

\begin{figure}[t!]
    \centering
    \includegraphics[trim=0cm 1.1cm 0cm 0.1cm, clip, width=0.4\textwidth]{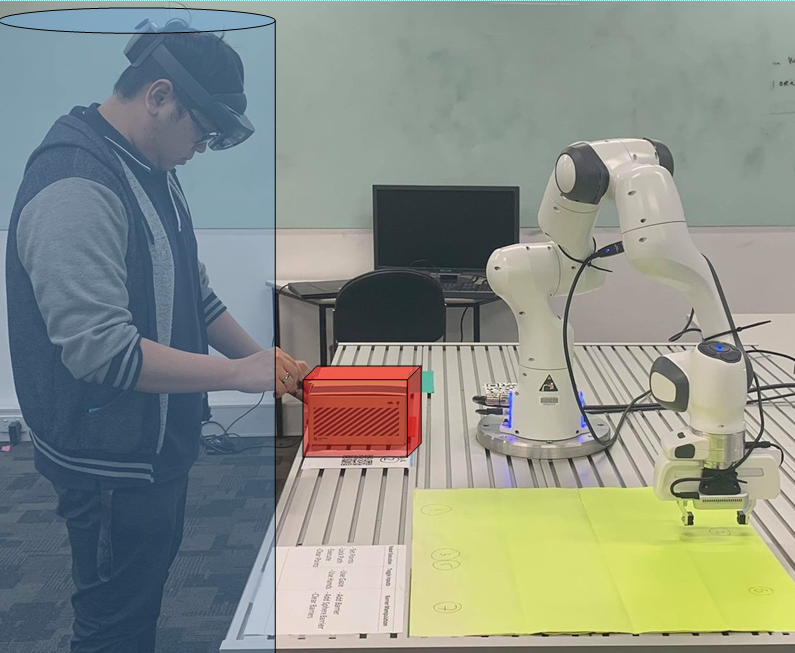}
    \caption{A user, wearing an augmented reality headset, is shown collaborating with the robot using our Virtual Barrier system. Using our system, users can collaborate with the robot while being protected by a Virtual Person Barrier (blue cylinder). Objects in the shared workspace can be protected by Virtual Obstacle Barriers (red rectangular prism).}
    \label{PersonBarrier}
\end{figure}

In recent years, augmented reality (AR) technology has been explored and shown to be a promising alternative to traditional programming interfaces with great potential for enabling fluent human-robot interaction \cite{Green,Pan}. With the ability to create visualizations co-located in the physical world, AR enables a rich and intuitive method for communicating task-related information to the user. Herein, we introduce a novel AR-based Virtual Barrier system for enabling safe and effective human-robot collaboration (Figure \ref{PersonBarrier}). Using our system, the user is protected during collaborative scenarios with two kinds of Virtual Barriers: 

\begin{itemize}
    \item \textit{Virtual Person Barrier}: a person-following cylindrical virtual geometry that surrounds the user wearing the AR headset. 

    \item \textit{Virtual Obstacle Barriers}: Our system allows creation and visualization of arbitrary virtual geometries in the workspace, utilizing the capability of AR in visualizing virtual objects co-located in the robot's physical workspace, and ability to interact with such virtual objects using intuitive hand gestures.
\end{itemize}

The Virtual Person Barrier protects the user from potential collisions with the robot. Virtual Obstacle Barriers can be created arbitrarily by the user to protect objects or regions in which collision with the robot must be avoided. After they are created, the Virtual Obstacle Barriers can be modified online via hand gestures. To facilitate seamless human-robot collaboration, our system also features a heads-up robot programming interface, as well as automatic online path re-planning whenever a virtual barrier interferes the robot's path (for example, when a user, surrounded by the Virtual Person Barrier, crosses a planned robot path). To evaluate our novel system, we conduct a user study with an experiment mimicking an industrial manufacture setting, where participants program robot trajectories and collaborate in the same workspace with the robot. Results show that participants feel safer working with the robot using our system and found our AR-based interface more intuitive than a standard 2D monitor-based robot programming interface. Results also indicate that our system increases task efficiency.

\section{RELATED WORKS}
        \label{sec:related_works}
        \subsection{Safety in Human Robot Interaction}
In order to prevent users from being harmed when working with robots, researchers have proposed various systems which emphasizes physical safety. Broquere et al. developed a planner where new trajectories are computed in real-time to prevent potential collisions from happening between human and robot \cite{Broquere}. Lafferanchi et al. focused on real-time control by keeping track of energy stored in the system and adjusting the trajectories to minimise contact force with user during collaborations involving physical contact \cite{laffranchi}. Flacco et al. utilizes depth sensors to evaluate the distance between robot and human to prevent potential collision in real-time \cite{flacco}.

In Lasota et al.'s survey of methods for safe HRI, the authors laid out the definition of safety and categorized it into two different forms: physical safety and psychological safety \cite{Lasota_book}. While physical harm can be prevented using methods presented in the papers mentioned above, this does not necessarily lead to stress-free and comfortable interaction. Arai et al. found that workers experience a greater subjective workload in the form of stress and anxiety due to close proximity with industrial robots \cite{Arai}. Lasota and Shah's study on close-proximity human-robot interaction \cite{lasota_shah} indicates that maintaining physical safety by simply avoiding impending collision can lead to low levels of perceived safety and comfort in humans. To ensure not only physical safety but also psychological safety, researchers have incorporated social considerations into their approaches. Mainprice and Berenson proposed a framework to facilitate safe collaboration between human and robot based on early prediction of the human's motion \cite{Mainprice_n_Berenson}. Sisbot and Alami developed a human-aware motion planner for mobile robots where human kinematics, vision fields, postures and preferences are taken into account for the robot's motion \cite{sisbot}. Another motion planner by Mainprice et al. uses HRI constraints to create safe robot motion, utilizing cost-based, random sampling method \cite{mainprice}.

\subsection{Augmented Reality in Human Robot Collaboration (HRC)}
Not knowing the robot partner's intent often contributes to stressful situations for the human user. Being able to convey robot's intent would greatly reduced the anxiety formed when working with robots, and Augmented Reality (AR) has proven its potential in addressing this issue. Brending et al. suggested that by utilising AR in HRC, anxiety could be reduced when contextual information is shown to the human operator working in close proximity with a robot \cite{brending}.

Recognizing this potential, several AR-based safety systems have been proposed over the last few years. Zaeh et al. proposed a robot programming interface for industrial robots utilising AR in which trajectories and target coordinates can be visualized and manipulated with interactive laser projections \cite{Zaeh}. Gaschler et al. proposed an AR-based system with virtual obstacles in an augmented workspace being manipulated and displayed on a 2D display to produce collision-free trajectories \cite{gaschler}. Michalos et al. developed an AR-based tool to support users while they are sharing the workspace with robots \cite{michalos}. However, these AR interfaces utilizing 2D displays or handheld devices require the user to shift their focus between the display and the physical workspace, and/or limit the availability of the user's hands to work on the collaborative task, reducing efficiency. While projector-based systems where virtual objects are overlaid on top of human-robot shared workspace are proposed as an alternative to resolve the aforementioned issue \cite{vogel,andersen2016projecting}, they suffer from occlusion problems. 

The use of AR head-mounted displays (i.e., AR headsets) address the aforementioned issues, and further allow for more fluent human-robot collaboration by incorporating natural input methods such as gestures and speech. Chan et al. developed a framework using an AR headset to perform robot programming using multimodal communications including gesture control and tactile feedback \cite{chan}. Rosen et al. proposed a system where controlling the robot arm and communicating its motion intent are done via a AR headset \cite{rosen}. Hietanen et al. developed a safety system where a depth sensor is used to create safety zones and display them using the AR headset \cite{hietanen}.

Based on the demonstrated potential for AR to address both physical and psychological safety, as well as task efficiency in HRC, this paper introduces a novel AR system that enables the creation, visualization, and manipulation of in-situ virtual barriers for protecting the user and the surroundings in real time. The organization of this paper is as follows. Section \ref{sec:system_architecture} describes the details of our system. Section \ref{sec:user_study} documents the user study for evaluating our system, with Sections \ref{sec:analysis_results} and \ref{sec:discussion} providing the results and discussion and Section \ref{sec:conclusion} concludes the paper. 

\section{SYSTEM ARCHITECTURE}
        \label{sec:system_architecture}
        \begin{figure}[t]
    \centering
    \includegraphics[width=0.42\textwidth,angle=270
    ,trim=1cm 3.6cm 3.5cm 9.5cm
    ,clip]{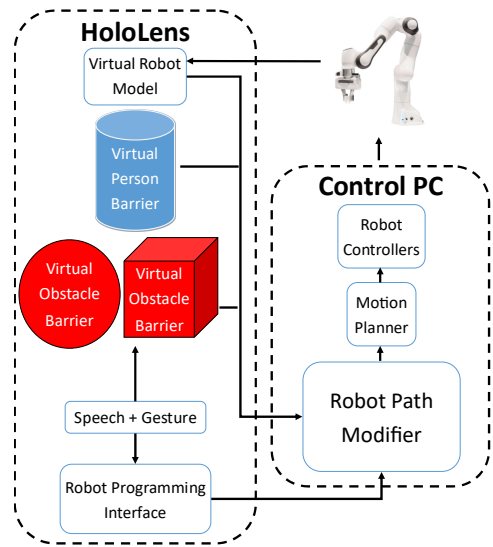}
    \caption{System diagram of our Virtual Barrier system}
    \label{main_comp}
\end{figure}

Our system consists of two main components. The AR interface implemented on a HoloLens 2 head-mounted AR display device\footnote{Microsoft HoloLens 2|https://www.microsoft.com/en-us/hololens}, and the robot motion planning and control system implemented using Robot Operating System (ROS) on a control PC (Figure \ref{main_comp}). The following subsections explain the modules in each component.
 
        \subsection{AR Interface}
        \label{subsec:hololens}
        \textbf{Virtual Robot Model:} We created a one-to-one scale virtual robot model co-located with the real robot. The virtual model is connected to the real robot and mimics the real robot's movements. An AR tag placed at a known location relative to the real robot is used to calibrate the AR system at startup. The virtual model is used to confirm the calibration and connectivity between the virtual and real robot/system.

\textbf{Virtual Person Barrier:} To protect the user when collaborating with the robot in the same physical workspace, we created a Virtual Person Barrier - a virtual cylinder geometry encapsulating the user, placed at the location of the HoloLens headset (blue cylinder in Figure \ref{PersonBarrier}). This barrier acts as a virtual obstacle that protects the user by following the user (headset) when they move around the workspace. To avoid distracting the user from performing their task, the person barrier is rendered with low opacity.

\begin{figure}[h]
    \centering
    \includegraphics[width=0.2\textwidth,angle=270
    ,trim=2.4cm 0cm 12cm 11.5cm
    ,clip]{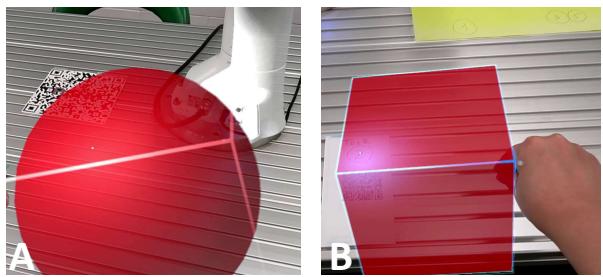}
    \caption{Manipulable Virtual Barriers rendered in AR by the HoloLens. Users can interact and modify the virtual barriers (including moving, scaling, and rotating) in realtime using intuitive hand gestures.}
    \label{Virtual_Barriers}
\end{figure}

\textbf{Virtual Obstacle Barriers:} Our system allows users to arbitrarily place geometries in AR to act as virtual barriers for protecting objects or regions in the workspace (as shown in Figure \ref{Virtual_Barriers}). Such virtual barriers can come in different shapes (sphere in \ref{Virtual_Barriers}A or cube in \ref{Virtual_Barriers}B), and can be spawned using gaze and a speech command, or can be dragged-and-dropped from a menu rendered in AR using hand gestures. After placing virtual obstacle barriers in the workspace, the user can also interact and manipulate them (reposition, scale, rotate) in real-time using intuitive hand gestures, for example pinching (as shown in Figure \ref{Virtual_Barriers}B).

\begin{figure}[h]
    \centering
    \includegraphics[width=0.3\textwidth]{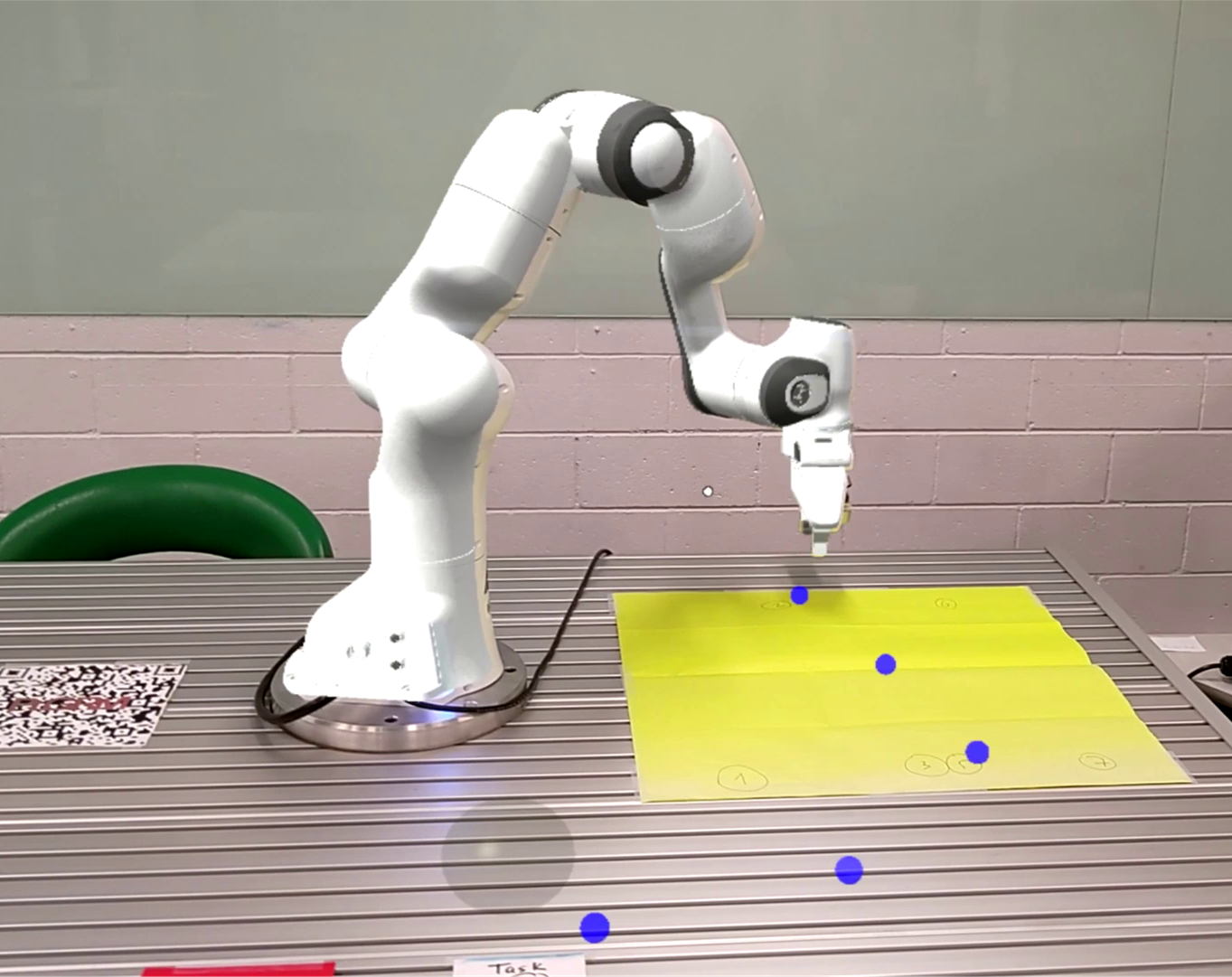}
    \caption{Visual robot programming interface for the Virtual Barrier system. Users set way-points (blue points) using head gaze along with speech command. This path can then be executed using another speech command.}
    \label{Robot_Programming}
\end{figure}

\textbf{Robot Programming Interface:} We provide a heads-up robot programming interface with our AR system, utilizing hand gestures and speech (as shown in Figure \ref{Robot_Programming}). A way-point programming interface is adapted from \cite{chan}, where users can provide desired robot end-effector way-points with either head gaze or a hand ray pointer. Multiple sub-paths can be created using a motion planner (provided by MoveIt Motion Planning Framework \cite{moveit}) to connect the way-points together. The system also accepts several voice commands to assist the user with robot motion planning such as ``\textbf{LOCK PATH}" to send the series of set way-points for motion planning and ``\textbf{EXECUTE}" to command the robot to execute the planned path.
        \subsection{Robot Motion Planning and Control}
        \label{subsec:ros}
        \textbf{Robot Path Modifier:} This module monitors the location of Virtual Barriers to ensure collision-free sub-paths are created between user-defined way-points. An automatic, online path re-planner, running at 0.75Hz is employed. When a sub-path is detected to be invalid, (i.e. intersecting with a Virtual Barrier such as a moving Virtual Person Barrier), the path is modified by the motion planner so that collisions are avoided. We implement two robot behaviors when potential collisions are detected.
For safety, if the robot is currently executing an invalid sub-path, it will temporarily stop and a new sub-path will be updated for the robot to execute. If the invalid sub-path(s) are not currently being executed by the robot, the future sub-path(s) are updated without stopping the robot. Sub-paths are generated using MoveIt's motion planner.
        
\section{USER STUDY}
        \label{sec:user_study}
\begin{figure}[h]
    \centering
    \includegraphics[trim=0.1cm 0cm 0cm 0cm, clip, width=0.32\textwidth]{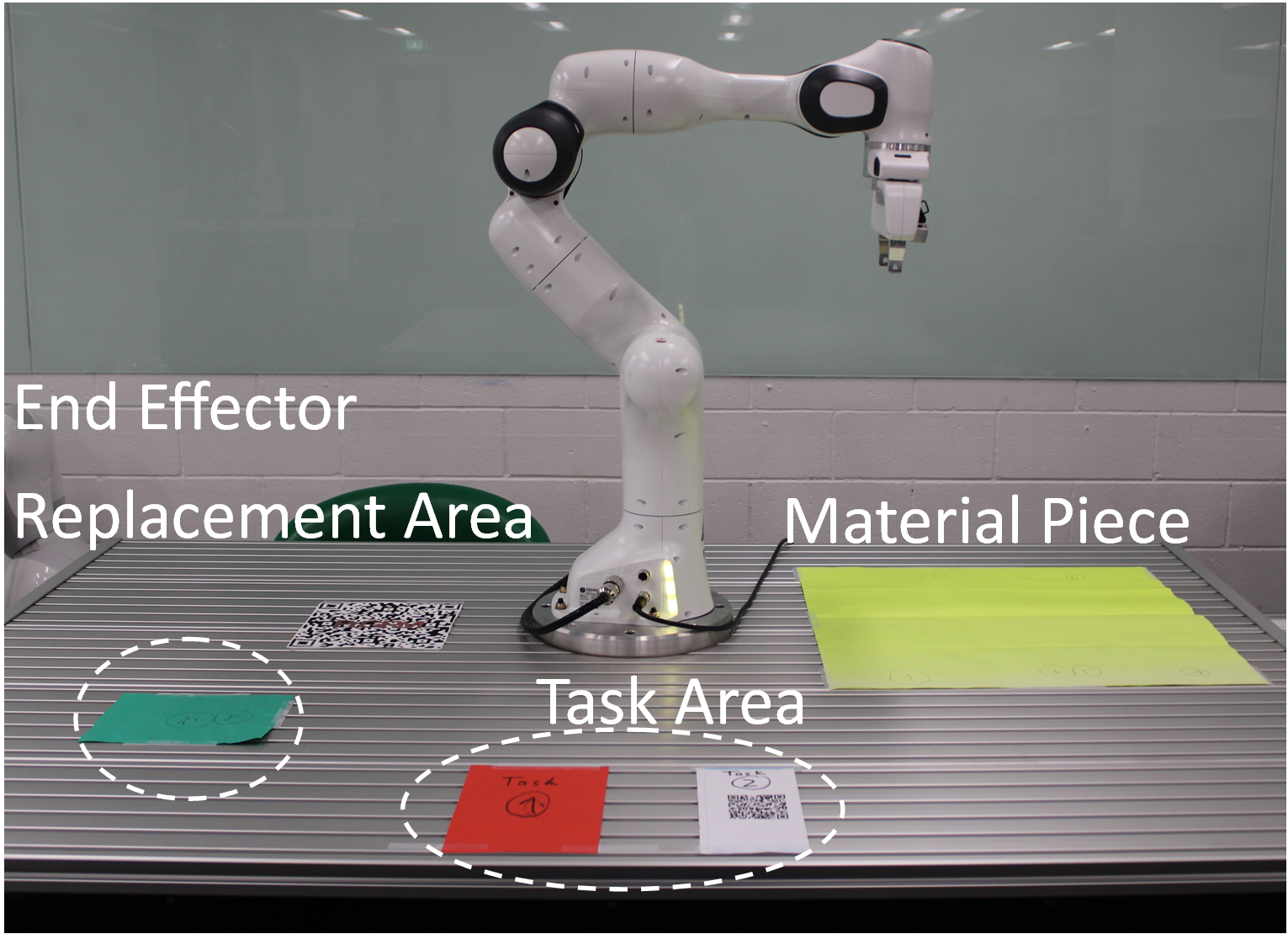}
    \caption{Experimental setup.}
    \label{experiment_setup}
\end{figure}

        We evaluated our system in an experiment, where participants are asked to perform a task mimicking a sheet material smoothing task (a routine procedure in industrial carbon-fibre-reinforced-polymer manufacturing process) with the help of the robot. This procedure exemplifies typical collaborative tasks, where both human robot need to operate in and share the same physical workspace. In this use case, the user instructs the robot to follow a specified path to smooth a sheet of material laid out on the workspace, while performing auxiliary tasks (e.g., inspection) in the same workspace as the robot is operating. The experiment setup is shown in Figure \ref{experiment_setup}. The experiment starts with robot in a home position as shown in Figure \ref{experiment_setup}, and the participant programming eight predefined way-points for the robot to move to. The first three way-points are in the ``Material Piece Area", the fourth way-point is in the ``End-Effector Replacement Area", and next three way-points are again in the ``Material Piece Area" and the last way-point is in the ''End-Effector Replacement Area". All way-points are labeled clearly on the workspace for the participant to see. 

After setting all the way-points, the participant then commands the robot to execute the planned path, and proceeds to perform two other tasks in the ``Task Area", as the robot starts moving. The first task is setting up (placing) a toolbox at a predefined location in the shared workspace and protecting it by placing a virtual obstacle barrier around it using the provided interface. The second task is performing an inspection task via capturing a QR code placed in the shared workspace with a handheld device. The experiment is concluded when both the robot finishes its smoothing task and the participant finishes their two tasks. During the task, whenever a collision occurs, or is about to occur, the robot motion is halted and restarted by the experimenter (for safety), before the task can resume. The time taken to redo path planning when the experiment is restarted will be taken into account as a penalty for collision.

\begin{figure}[h]
    \centering
    \vspace{-6mm}
    \includegraphics[width=0.40\textwidth,angle=270
    ,trim=0cm 0.4cm 3.9cm 6.5cm
    ,clip]{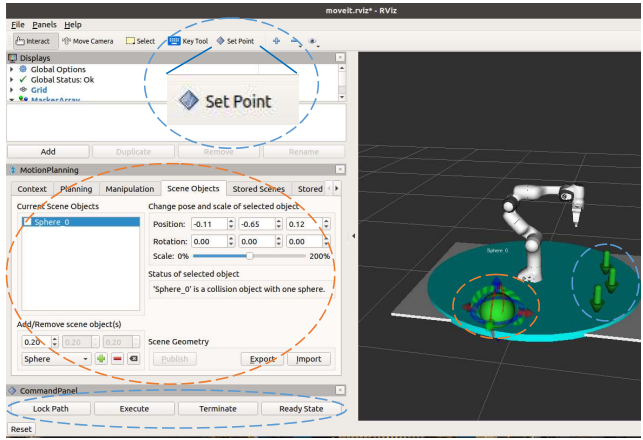}
    \vspace{-8mm}
    \caption{Robot Programming Interface on 2D display. Waypoint Path Programming can be done the Rviz tools and panel being circled with blue dotted lines. Users can also place collision geometries with the tool provided by MoveIt (circled with orange dotted line).}
    \vspace{-3mm}
    \label{computer}
\end{figure}

We compare our novel Virtual Barrier system with a standard robot programming interface on 2D display (as shown in Figure \ref{computer}), representative of the current standard of computer terminal and teach pendant-based interfaces used in the manufacturing industry. The interface provides a 3D view of the robot in the shared workspace, using Rviz. Rviz is also used to provide a point-and-click way-point placing tool in the 3D view for robot path programming. The standard MoveIt \cite{moveit} Rviz plugin is used to provided the functionality of placing collision geometries into the 3D view for protecting objects in the workspace. Both Rviz and MoveIt are widely used interfaces for robot control and interaction.

Participants are asked to perform the experiment task with the two interfaces. We counterbalanced the ordering in which the two interfaces are presented to the participants. Prior to each iteration of experiment, there is a training phase for the interface the participants would use for that iteration. The training phase includes an introductory video to the interface, as well as a 5-minute practice session. At the end of each iteration, experience with each interface are surveyed with a questionnaire. There are 7 items in the questionnaire and the participants are asked to give their responses on a 7-point Likert scale, with 1 indicating Strongly Disagree and 7 indicates Strongly Agree. Table \ref{tab:my-table} shows the list of questions. At the end of the study, the participants are also asked to fill in a feedback form about their preferred interface, overall experience, and thoughts on how the Virtual Barrier system with AR headset could be improved. Approval from the Monash University Human Research Ethics Committee was obtained prior to commencing the study (ethics application ID: 24867).

\begin{table*}[]
\centering
\caption{Questions asked in the survey after each iteration}
\label{tab:my-table}
\begin{tabular}{p{3.5cm}p{12cm}}
\hline
Categories      &Individual Questions \\ \hline
Safety          & Q1. I feel collisions between me and the robot can happen anytime.\\
                & Q2. I feel confident that the robot would not damage the surroundings (including objects in the workspace).\\
                & Q3. I feel confident in my programming and interacting with the robot using this method.\\
Intuitiveness   & Q4. A lot of time is required to learn the interface in order to do the programming task efficiently.\\
                & Q5. The interface intuitively helps me completing the overall task (i.e robot programming \& interacting).\\
Efficiency      & Q6. The interface helps me complete the task efficiently.\\
Preference      & Q7. I would prefer using this system for future tasks.\\ \hline
\end{tabular}
\end{table*}

We formulate the following hypotheses for our experiment. 

\textbf{Hypothesis 1:} The Virtual Barrier system will make human robot collaboration more \textbf{intuitive}.

\textbf{Hypothesis 2:} The Virtual Barrier system will increase task \textbf{efficiency} by decreasing the \textbf{task completion time}.

\textbf{Hypothesis 3:} The Virtual Barrier system will increase \textbf{user’s sense of safety} when interacting
with the robot, in terms of both \textbf{safety of the user} and  \textbf{safety of the robot}.

\section{ANALYSIS \& RESULTS}
        \label{sec:analysis_results}
\begin{figure}[h]
    \centering
    \includegraphics[width=0.4\textwidth
    ,trim=2.35cm 9cm 2.35cm 2cm
    ,clip]{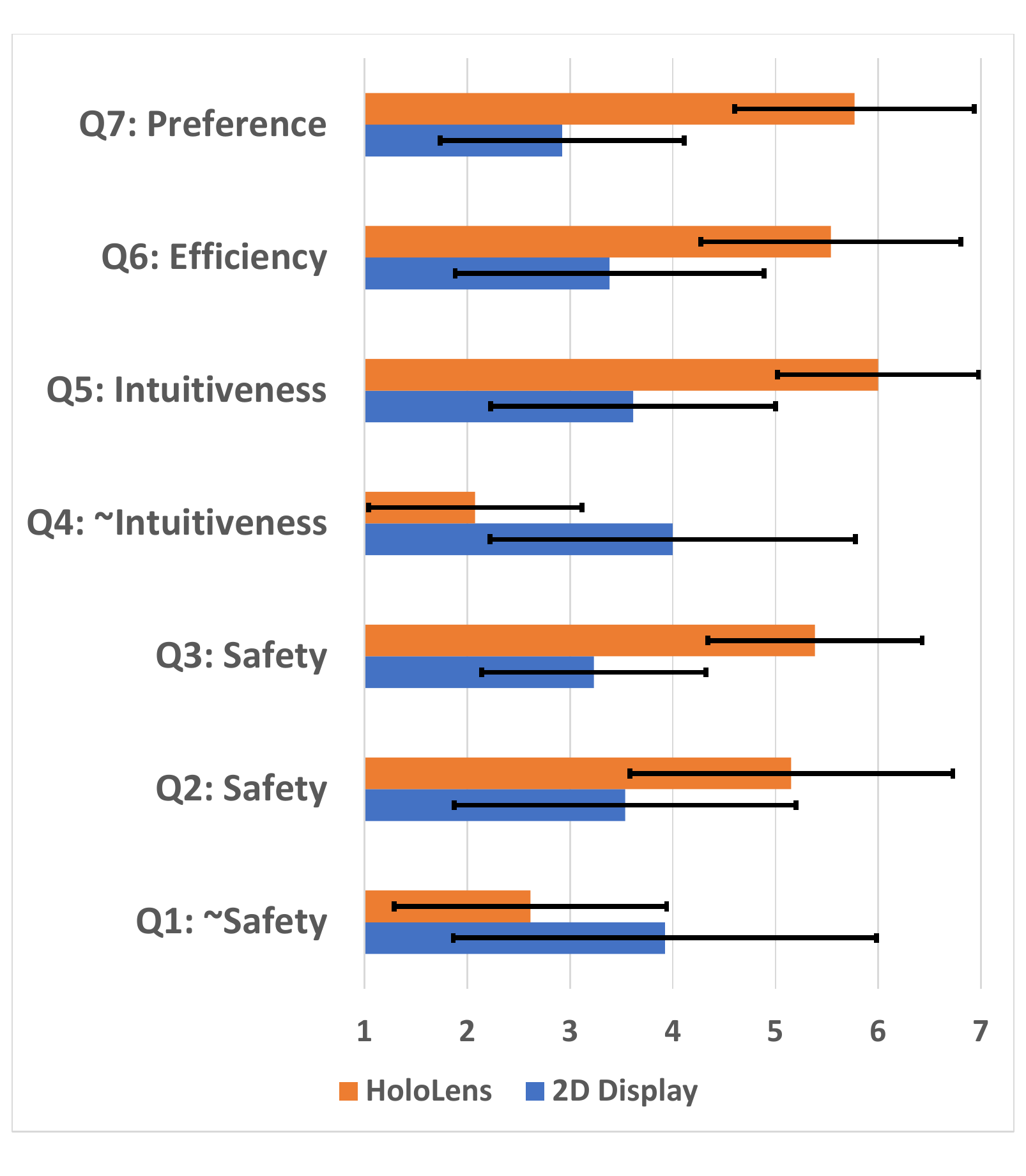}
    \vspace{-3mm}
    \caption{Mean scores from the survey (7-point Likert scale). Labels preceded by a tilde (\texttildelow) indicates that lower score means better performance for the questionnaire item.}
    \label{response}
    \vspace{-4mm}
\end{figure}

\begin{figure}[h]
    \centering
\includegraphics[width=0.4\textwidth
    ,trim=2.35cm 22cm 2.35cm 2cm
    ,clip]{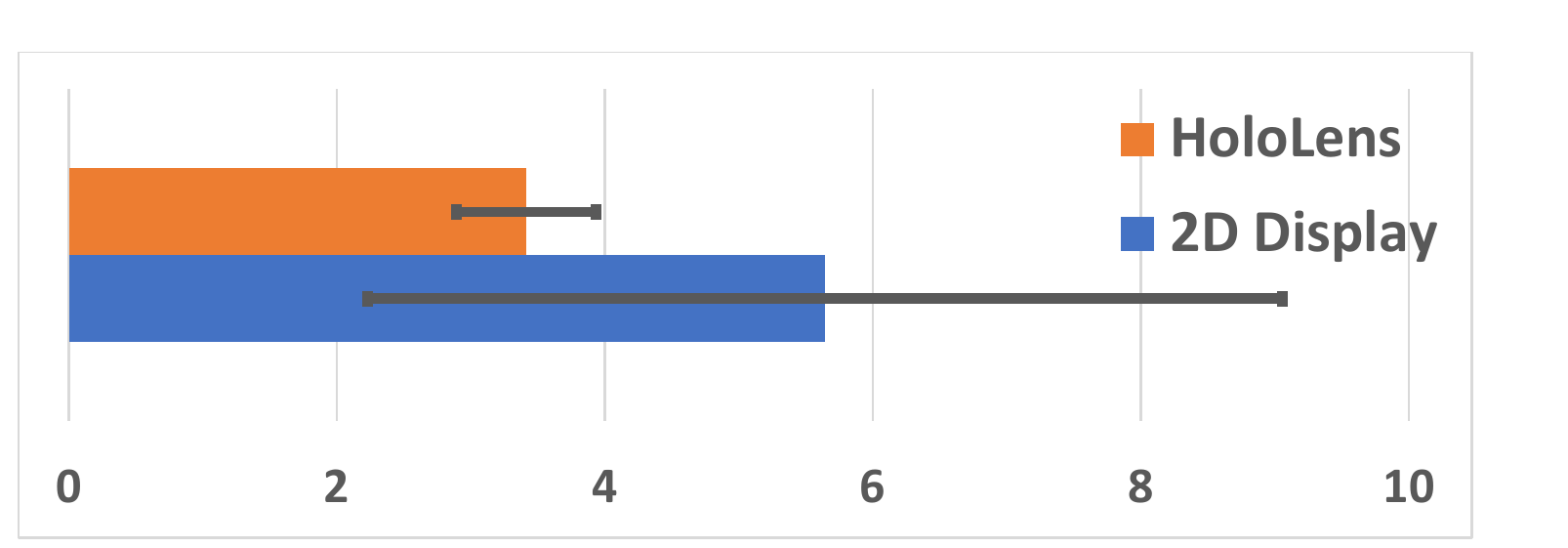}
    \caption{Mean Task Completion Time in minutes from the user studies}
    \label{task_completion_time}
    \vspace{-2mm}
\end{figure}

        Our study included 13 participants (11 males and 2 females). There are 6 participants out of 13 that did not have any experience with robot programming, and there are 8 out of 13 that did not having any experience with AR. The mean and standard deviation of the scores for the questions are shown in Figure \ref{response}. The task completion time for each condition is also shown in Figure \ref{task_completion_time}.
Paired-t test was used for hypothesis testing. Significance are reported at $\alpha$ = 0.05. 

Results show that our Virtual Barrier system provides a better experience in terms of safety (Q1, t(12) = 2.85, p = 0.007, Q2, t(12) = -3.41, p = 0.002) intuitiveness (Q3, t(12) = -6.79, p $<$ 0.0005, Q4, t(12) = 4.06, p = 0.0007, Q5, t(12) = -6.2, p $<$ 0.0005) and efficiency (Q6, t(12) = -6.06, p $<$ 0.0005, task completion time, t(12) = 6.54, p $<$ 0.0005). In terms of preference, participants much prefer our system over the standard 2D display interface for future tasks (Q7, t(12) = -7.01, p $<$ 0.0005).

\section{DISCUSSION}
        \label{sec:discussion}
        \subsection{Psychological Safety}
\label{psy_safe}
The results of the survey (Questions 1-3) support \textbf{Hypothesis 3}, that the proposed system will increase user's sense of safety. However, some responses received from the users have piqued our interest. In the response from Question 1 in the survey, the Virtual Barrier system outperforms the 2D display interface overall, as participants on average reported that they feel collision with the robot is less likely to happen with the Virtual Barrier system. However, inspecting the individual responses, about half of the participants  gave a slightly worse rating for the Virtual Barrier system. The close rating for sense of safety (collision with robot) between the two interfaces perhaps suggests that, even though there was no protection from the 2D display-based collaboration interface to prevent collision, participants were simply unaware of the potential danger when working with a robot. This correlates with the lack of experience with robots reported by 6 out of 13 participants, and the fact that the robot we used for the experiment is a robotic arm designed specifically for HRC, with a friendlier appearance such as white colour, curved contours, and slower movements, as opposed to typical industrial robots, where their appearance are more intimidating to users. 

Furthermore, for the HoloLens condition, although the majority of participants indicated that they disagree (giving a response of less than 4) with Question 1, which states collision with robot can happen anytime, some participants agreed with the statement (giving a response more than 4), commenting that ``\textit{...the robot would hit me because I need to physically be close to the robot...}", even though the participant is protected by the Person Barrier with the Virtual Barrier system. This might be a result of the Person Barrier being rendered too transparent (to avoid creating distraction from the main task), and participants not seeing it very definitely. However, the benefits of the Virtual Barrier system in terms of safety are clearly reflected by the quantitative results and comments from other participants, stating that ``\textit{...I felt much safer using the HoloLens, as the arm made an effort to avoid the user...}" and ``\textit{... it was a good experience compared with the computer interface...}".

\subsection{Physical Safety}
Aside from the data presented in Section \ref{sec:analysis_results}, we counted the number of times that the robot had to be stopped by the experimenter in each condition (due to collision/near collision with participant/surroundings). With our AR system, there were zero occasions where the robot needed to be stopped by the experimenter, as the Person Barrier and online replanning mechanism (mentioned in Section \ref{subsec:hololens}) enables the robot to automatically avoid collisions with the person. However, with the standard 2D monitor interface, the robot needed to be stopped manually about 1-2 times per participants on average. As most participants were inexperienced with robots, they were not aware of the potential danger when working with the robot, thus would sometimes position themselves where collisions with robot would occur. These observations demonstrate our systems benefit in ensuring physical safety in HRC.

\subsection{Intuitiveness}
\label{intuitiveness}
The results from the survey (Question 4 and 5) supported \textbf{Hypothesis 1}; namely that the Virtual Barrier system makes the interaction between human and robot more intuitive. However, participant comments revealed some difficulties in using the AR system and pointed towards potential improvements that can be made, including ``\textit{...having better depth perception...}" and ``\textit{...mak[ing] boxes easier to manipulate if possible...}". As most participants reported no prior experience with AR, it is evident that the steep learning curve of using AR devices is still a problem, despite receiving multiple forms of induction with the AR device via introductory video as well as practice session prior to the experiment. Based on these findings, we recommend that longer periods of training are given when introducing AR-base interfaces to unfamiliar workers, in combination with some form of evaluation to gauge level of comfort with the technology.

\subsection{Efficiency}
The results from the user study supports \textbf{Hypothesis 2}, with both results from Question 6 in the survey and the average task completion time favouring the proposed AR interface. A large standard deviation in task completion time for the 2D display interface was observed with completion time ranging between 3 minutes 40 seconds and 8 minutes. Many participants struggled to perform robot programming task using the 2D display interface. This might be caused by the difference in perspective between the real world and the 2D display. Programmed way-points were sometimes either placed too close to the robot, which caused the robot to enter self-collision, or too far away out of the robot's reach. These resulted in multiple attempts required in order to complete the task. Moreover, while using the 2D display to place collision geometry, users generated incorrect placements, which resulted in the robot colliding with objects placed in the shared workspace. The robot had to be hand-guided out of the collision state by the experimenter when the situation occurred, which resulted in a greatly increased time taken to finish the task under the 2D display condition.

\subsection{Preference}
The average score of Question 7 in the survey shows that our Virtual Barrier system is preferred by participants when compared to a standard 2D display-based interface for performing human-robot collaborative task. Also in the general feedback at the end, all 13 participants indicated that their preferred method of robot programming is with the AR interface. This suggests the idea that despite the steep learning curve associated with AR as mentioned in Section \ref{intuitiveness}, AR-based interfaces such as our Virtual Barrier system still has great potential in providing safe, intuitive, efficient robot programming interfaces.
        
\section{CONCLUSION AND FUTURE WORK}
        \label{sec:conclusion}
        In this paper, we introduced a novel AR-based system for facilitating safe and effective human-robot collaboration. With the use of our novel Virtual Barrier system, user and surrounding environment are protected from collision with the robot by virtual geometries co-located in the physical workspace, visualized via AR. We conducted a user study to investigate the benefits that the system could bring, and results showed that our Virtual Barrier system improves safety, intuitiveness and effectiveness of HRC compared to a standard 2D display interface. Despite the improvements, many participants have commented on some challenges when using the system, especially when perceiving and interacting with virtual objects in AR due to the steep learning curve of the technology. 

In future work, based on our study results, we plan to investigate more intuitive methods of interacting with virtual objects, as well as the benefits longer training and familiarization periods provide for novice users.  We will also broaden our study to consider a wider range of users.

\bibliographystyle{IEEEtran}
\bibliography{ref}

\end{document}